\documentclass[11pt,a4paper]{article}
\usepackage[hyperref]{naaclhlt2019}
\usepackage{times}
\usepackage{latexsym}

\usepackage{url}
\usepackage{graphicx}
\usepackage{caption}
\usepackage{subcaption}
\usepackage{textgreek}
\usepackage{booktabs}
\usepackage{multirow}
\usepackage{todonotes}
\usepackage{array}
\usepackage{textcomp}

\graphicspath{ {./images/} }
\aclfinalcopy 
\def\aclpaperid{1721} 

\newcolumntype{s}{>{\small}c}
\newcolumntype{k}{>{\small}l}
\newcolumntype{j}{>{\small}r}

\title{Correlation Coefficients and Semantic Textual Similarity}

\author{Vitalii Zhelezniak, Aleksandar Savkov, April Shen \& Nils Y. Hammerla \\
Babylon Health \\
\texttt{\{firstname.lastname\}}
\texttt{@babylonhealth.com}
}

\date{}

\usepackage{amsmath}
\usepackage{amssymb}
\newcommand{\mathbold}[1]{\ensuremath{\boldsymbol{\mathbf{#1}}}}

\newcommand{\mbw}{\mathbold{w}}
\newcommand{\mbx}{\mathbold{x}}
\newcommand{\mby}{\mathbold{y}}

\newcommand{\mbW}{\mathbold{W}}

\usepackage{cleveref}  

\begin{document}
\maketitle
\begin{abstract}
A large body of research into semantic textual similarity has focused on constructing state-of-the-art embeddings using sophisticated modelling, careful choice of learning signals and many clever tricks.
By contrast, little attention has been devoted to similarity measures between these embeddings, with cosine similarity being used unquestionably in the majority of cases.
In this work, we illustrate that for all common word vectors, cosine similarity is essentially equivalent to the Pearson correlation coefficient, which provides some justification for its use.
We thoroughly characterise cases where Pearson correlation (and thus cosine similarity) is unfit as similarity measure.
Importantly, we show that Pearson correlation is appropriate for some word vectors but not others.
When it is not appropriate, we illustrate how common non-parametric rank correlation coefficients can be used instead to significantly improve performance.
We support our analysis with a series of evaluations on word-level and sentence-level semantic textual similarity benchmarks.
On the latter, we show that even the simplest averaged word vectors compared by rank correlation easily rival the strongest deep representations compared by cosine similarity.
\end{abstract}

\section{Introduction}
\label{sec:intro}

Textual embeddings are immensely popular because they help us reason about the abstract and fuzzy notion of semantic similarity in purely geometric terms.
Distributed representations of words in particular \citep{Bengio2003, Mikolov2013, Pennington2014, Bojanowski2016, Joulin2016a} have had a massive impact on machine learning (ML), natural language processing (NLP),
and information retrieval (IR).

Recently, much effort has also been directed towards learning representations for larger pieces of text,
with methods ranging from clever compositions of word embeddings \citep{Mitchell2008, DeBoom2016, Arora2017, Wieting2016, Wieting2018, zhelezniak2019}
to sophisticated neural architectures \citep{Le2014, Kiros2015, Hill2016,  Conneau2017, Gan2017, Tang2017, Zhelezniak2018, Subramanian2018, Pagliardini2018, USE2018}.

Comparatively, there is little research into similarity measures for textual embeddings.
Despite some investigations into alternatives \citep{Camacho-Collados2015, DeBoom2015, Santus2018, zhelezniak2019}, cosine similarity has persistently remained the default and unquestioned choice across the field.
This is partly because cosine similarity is very convenient and easy to understand.
Sometimes, however, we have to resist what is convenient and instead use what is appropriate.
The core idea behind our work is to treat each word or sentence embedding as a sample of (e.g. 300) observations from some scalar random variable.
Hence, no matter how mysterious word vectors appear to be, just like any samples, they become subject to the full power of traditional statistical analysis.
We first show that in practice, the widely used cosine similarity is nothing but the Pearson correlation coefficient computed from the paired sample.
However, Pearson's $r$ is extremely sensitive to even slight departures from normality, where a single outlier can conceal the underlying association.
For example, we find that Pearson's $r$ (and thus cosine similarity) is acceptable for word2vec and fastText but not for GloVe embeddings.
Perhaps surprisingly, when we average word vectors to represent sentences, cosine similarity remains acceptable for word2vec, but not for fastText any longer.
We show that this seemingly counterintuitive behaviour can be predicted by elementary univariate statistics, something that is already well known to researchers and practitioners alike.
Furthermore, when there are clear indications against cosine similarity, we propose to repurpose rank-based correlation coefficients, such as Spearman's $\rho$ and Kendall's $\tau$, as \emph{similarity measures} between textual embeddings.
We support this proposition by a series of experiments on word- and sentence-level semantic textual similarity (STS) tasks.
Our results confirm that rank-based correlation coefficients are much more effective when the majority of vectors break the assumptions of normality.
Moreover, we show how even the simplest sentence embeddings (such as averaged word vectors) compared by rank correlation easily rival recent deep representations compared by cosine similarity.

\section{Related Work}
\label{sec:related}

At the heart of our work is a simple statistical analysis of pre-trained word embeddings and exploration of various correlation coefficients as proxies for semantic textual similarity.
Hence, any research that combines word embeddings with tools from probability and statistics is relevant.
Of course, word embeddings themselves are typically obtained as the learned parameters of statistical machine learning models.
These models can be trained on large corpora of text to predict a word from its context or vice versa \citep{Mikolov2013}.
Alternatively, there are also supervised approaches \citep{Wieting2015,Wieting2016,Wieting2017,Wieting2018}.

A different line of research tries to move away from learning word embeddings as point estimates and instead model words as parametric densities \citep{Vilnis2014,Barkan2016,Athiwaratkun2017}.
These approaches are quite appealing because they incorporate semantic uncertainty directly into the representations.
Of course, such representations need to be learned explicitly.
In some cases one could estimate the densities even for off-the-shelf embeddings,
but this still requires access to the training data and the usefulness of such post-factum densities is limited \citep{Vilnis2014}.
In other words, these approaches are not very helpful to practitioners who are accustomed to using high-quality pre-trained word embeddings directly.

Arguably, statistical analysis of pre-trained word embeddings is not as principled as applying a probabilistic treatment end-to-end.
Any such analysis, however, is very valuable as it provides insights and justifications for methods that are already in widespread use.
For example, removing the common mean vector and a few top principal components makes embeddings even stronger and is now a common practice~\cite{Mu2018allbutthetop, Arora2016,Arora2017,Ethayarajh2018}.
These works view word embeddings as observations from some $D$-dimensional distribution; such treatment is naturally suitable for studying the overall geometry of the embedding space.
We, on the other hand, are interested in studying the similarities between individual word vectors and require a completely different perspective.
To this end, we see each word embedding itself as a sample of $D$ observations from a scalar random variable.
It is precisely this shift in perspective that allows us to reason about semantic similarity in terms of correlations between random variables and make the connection to the widely used cosine similarity.

Finally, we propose using rank-based correlation coefficients when cosine similarity is not appropriate.
Recently,~\citet{Santus2018} introduced a rank-based similarity measure for word embeddings, called APSynP, and demonstrated its efficacy on outlier detection tasks.
However, the results on the word-level similarity benchmarks were mixed, which, interestingly enough, could have been predicted in advance by our analysis.

\section{Correlation Coefficients and Semantic Similarity}
\label{sec:theory}

\begin{figure*}[th!]
	\centering
	\includegraphics[width=0.8\textwidth]{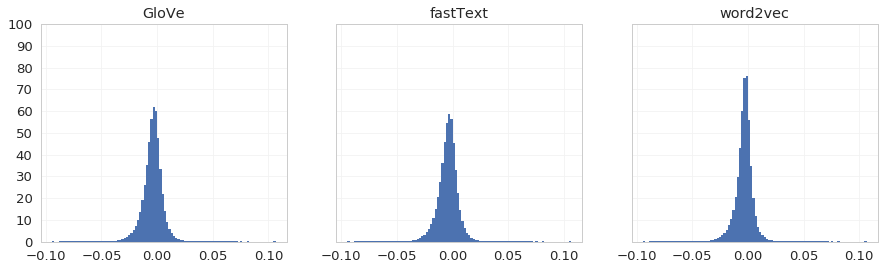}
	\caption{
	Normalised histograms of the mean distribution for three commonly used word embedding models:
	GloVe~\cite{Pennington2014}, fastText~\cite{Bojanowski2016}, and word2vec~\cite{Mikolov2013a,Mikolov2013b}.
	}
	\label{fig:histogram}
\end{figure*}

Suppose we have a vocabulary of $N$ words $\mathcal{V} = \{w_1, w_2, \ldots, w_N\}$ and the word embeddings matrix $\mbW \in \mathbb{R}^{N\times D}$, where
each row $\mbw^{(i)}$ for $i=1,\ldots,N$ is a $D$-dimensional word vector. Popular pre-trained embeddings in practice typically have dimension $D=300$, while the vocabulary size $N$ can range from thousands to millions of words.

We now consider the following: what kinds of statistical analyses can we apply to $\mbW$ in order to model semantic similarity between words?
One option is to view all word embeddings $\mbw^{(1)}, \mbw^{(2)}, \ldots \mbw^{(N)}$ as a sample of $N$ observations from some $D$-variate distribution $P(E_1, \ldots E_{D})$.
For example, we can fit a Gaussian and study how all $300$ dimensions correlate with each other.
Perhaps we can fit a mixture model and see how the embeddings cluster.
We could also normalise them and study their distribution on the unit sphere.
It is clear by now that $P(E_1, \ldots, E_{D})$ is suitable for describing the overall geometry of the embedding space but is not very useful for our goals.

If we are to reason about similarities between individual word vectors, we should instead be looking at the transpose of $\mbW$.
Putting it differently, we see $\mbW^{T}$ as a sample of $D$ observations from an $N$-variate distribution $P(W_1, W_2, \ldots, W_N)$,
where $W_i$ is a scalar random variable corresponding to the word $w_i$.
This distribution is exactly what we need because the associations between $W_i$ captured by $P$ will become a proxy for semantic similarity.
Often we are only interested in pairwise similarities between two given words $w_i$ and $w_j$; thus the main object of our study is the bivariate marginal $P(W_i, W_j)$.
To lighten up the notation slightly, we denote the two words as $w_x$ and $w_y$, and the corresponding random variables as $X$ and $Y$.
We also refer to $P(X, Y)$ as the joint and $P(X)$, $P(Y)$ as the marginals.
In practice, of course, the actual $P(X, Y)$ is unknown but we can make inferences about it based on our sample $(\mbx, \mby) = \{(x_1, y_1), (x_2, y_2), \ldots (x_{D}, y_{D})\}$.

First, we might want to study the degree of linear association between $X$ and $Y$, so we compute the sample Pearson correlation coefficient
\begin{equation}
\label{eq:pearson}
\hat{r} = \frac{\sum_{i=1}^{D}(x_i - \bar{x})(y_i - \bar{y})}{\sqrt{\sum_{i=1}^{D}(x_i - \bar{x})^2} \sqrt{\sum_{i=1}^{D}(y_i - \bar{y})^2}},
\end{equation}
where $\bar{x}$ and $\bar{y}$ are the sample means
\begin{align}
\label{eq:smean}
\bar{x} &= \sum_{i=1}^{D}x_i, & \bar{y} &= \sum_{i=1}^{D}y_i.
\end{align}
Let's view $\mbx$ and $\mby$ as word embeddings momentarily and compute cosine similarity between them
\begin{equation}
\label{eq:cosine}
\text{cos}(\mbx, \mby) = \frac{\sum_{i=1}^{D}x_i y_i}{\sqrt{\sum_{i=1}^{D}x_i^2} \sqrt{\sum_{i=1}^{D}y_i^2}}.
\end{equation}
We see now that~\Cref{eq:pearson} and~\Cref{eq:cosine} look very similar; when the sample means $\bar{x}$, $\bar{y}$ are zero, cosine similarity and Pearson's $\hat{r}$ are equal.
The real question here is whether or not they coincide in practice.
Putting it differently, if we take any single word vector $\mbw$ and compute the mean (across the $D$ dimensions), is this mean close to zero?
It turns out that it is, and we can show this by plotting the distribution of the means across the whole vocabulary for various popular word embeddings (see \Cref{fig:histogram}).
We find that the means are indeed highly concentrated around zero; quantitatively, only 0.03\% of them are above 0.05 in magnitude.
It follows that in practice when we compute cosine similarity between word vectors, we are actually computing Pearson correlation between them.

\begin{figure*}[!ht]
	\centering

	\begin{subfigure}{.3\textwidth}
		\centering
		\includegraphics[width=\textwidth]{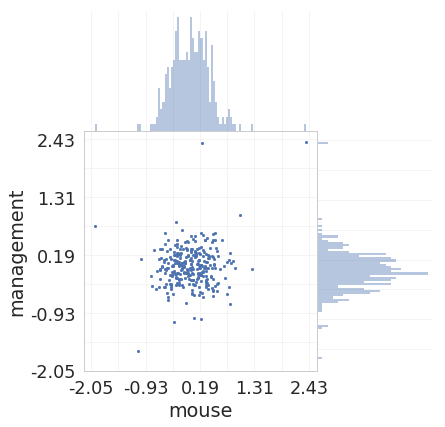}
	\end{subfigure}
	\begin{subfigure}{.3\textwidth}
		\centering
		\includegraphics[width=\textwidth]{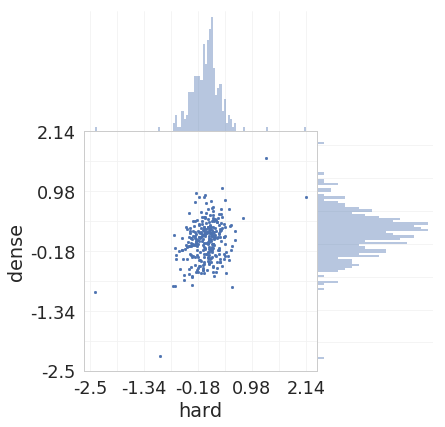}
	\end{subfigure}
	\begin{subfigure}{.3\textwidth}
		\centering
		\includegraphics[width=\textwidth]{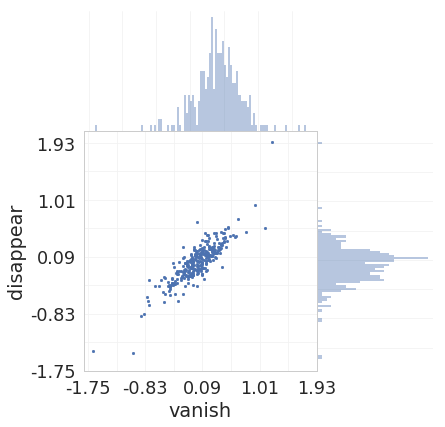}
	\end{subfigure}

	\begin{subfigure}{.3\textwidth}
		\centering
		\includegraphics[width=\textwidth]{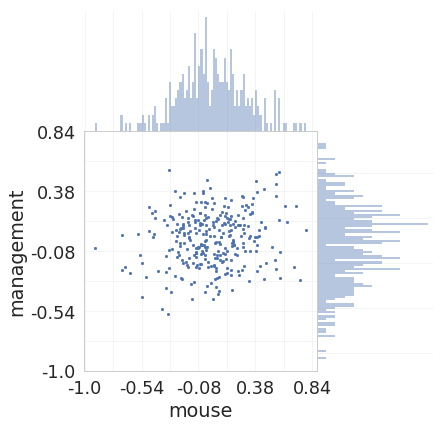}
	\end{subfigure}
	\begin{subfigure}{.3\textwidth}
		\centering
		\includegraphics[width=\textwidth]{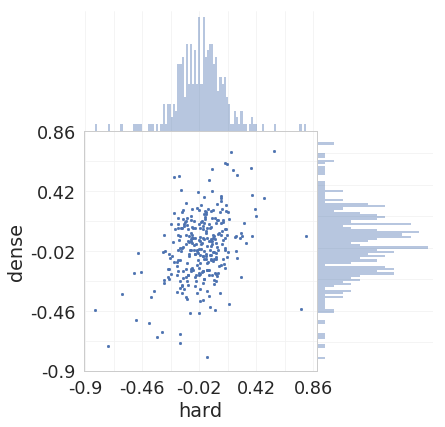}
	\end{subfigure}
	\begin{subfigure}{.3\textwidth}
		\centering
		\includegraphics[width=\textwidth]{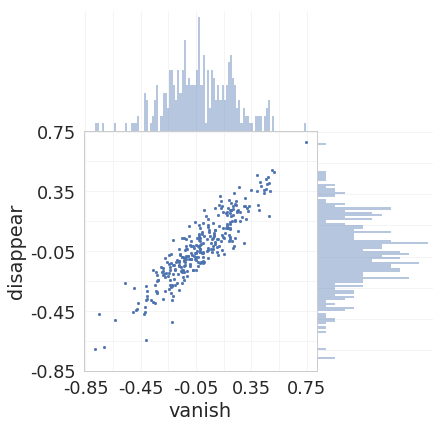}
	\end{subfigure}

	\begin{subfigure}{.3\textwidth}
		\centering
		\includegraphics[width=\textwidth]{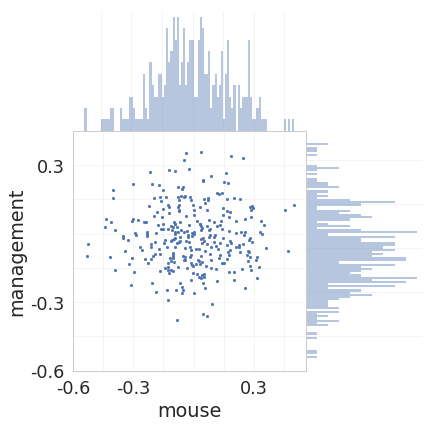}
	\end{subfigure}
	\begin{subfigure}{.3\textwidth}
		\centering
		\includegraphics[width=\textwidth]{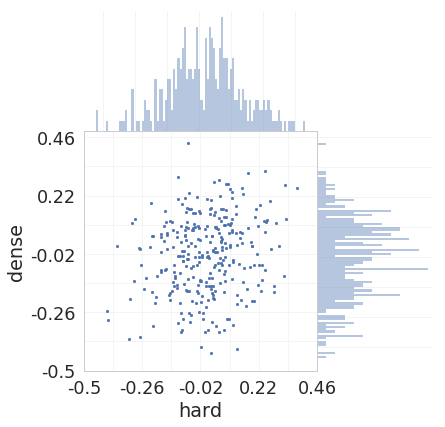}
	\end{subfigure}
	\begin{subfigure}{.3\textwidth}
		\centering
		\includegraphics[width=\textwidth]{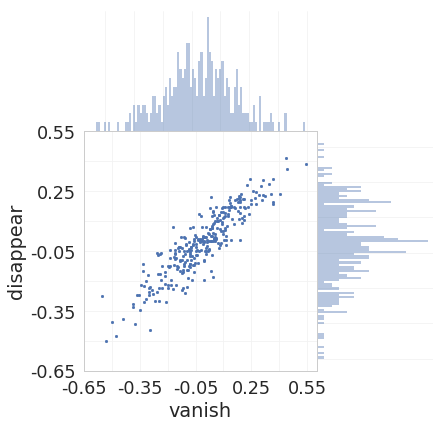}
	\end{subfigure}

	\caption{
	Scatter plots of paired word vectors, along with histograms (100 bins) of individual word vectors.
	Rows from top to bottom correspond to one of three common models: GloVe~\cite{Pennington2014}, fastText~\cite{Bojanowski2016}, and word2vec~\cite{Mikolov2013a,Mikolov2013b}.
	Columns from left to right correspond to increasing degrees of semantic similarity between the words, and accordingly increasingly pronounced
	linear correlation between the word vectors.
	Both the scatter plots and the histograms exhibit the presence of heavy outliers for GloVe vectors, which damage the efficacy of Pearson correlation in reliably capturing statistical associations.
	The outliers are relatively less pronounced for fastText vectors and much less pronounced for word2vec vectors.
	}

	\label{fig:scatter}
\end{figure*}

However, is this always the right thing to do? When the joint $P(X, Y)$ is bivariate normal, Pearson correlation indeed provides a complete summary of association between $X$ and $Y$,
simply because the covariance is given by $\text{cov}(X, Y)~=~r_{XY}\sigma_X\sigma_Y$.
However, Pearson correlation is extremely sensitive to even the slightest departures from normality -- a single outlier can easily conceal the underlying association \citep{Pernet2013}.
When the normality of $P(X, Y)$ is in doubt, it is preferable to use robust correlation coefficients such as Spearman's $\hat{\rho}$ or Kendall's $\hat{\tau}$.

Spearman's $\hat{\rho}$ is just a Pearson's $\hat{r}$ between ranked variables
\begin{equation}
\label{eq:spearman}
\hat{\rho} = \frac{\sum_{i=1}^{D}(r[x_i] - \overline{r[x]})(r[y_i] - \overline{r[y]})}{\sqrt{\sum_{i=1}^{D}(r[x_i] - \overline{r[x]})^2} \sqrt{\sum_{i=1}^{D}(r[y_i] - \overline{r[y]})^2}},
\end{equation}
where $r[x_i]$ denotes the integer rank of $x_i$ in a vector $\mbx$ (similarly $r[y_i]$), while $\overline{r[x]}$ and $\overline{r[y]}$ denote the means of the ranks.
Kendall's $\hat{\tau}$ is given by
\begin{equation}
\label{eq:kendall}
\hat{\tau} = \frac{2}{D(D-1)} \sum_{i < j}\text{sgn}(x_i - x_j)\text{sgn}(y_i - y_j)
\end{equation}
and can be interpreted as a normalised difference between the number of concordant pairs and the number of discordant pairs.
These rank correlation coefficients are more robust to outliers than Pearson's $\hat{r}$ because they limit the effect of outliers to their ranks: no matter how far the outlier is, its rank cannot exceed $D$ or fall below 1 in our case.
There are also straightforward extensions to account for the ties in the ranks.

\begin{figure*}[!ht]
	\centering
	\includegraphics[width=\textwidth]{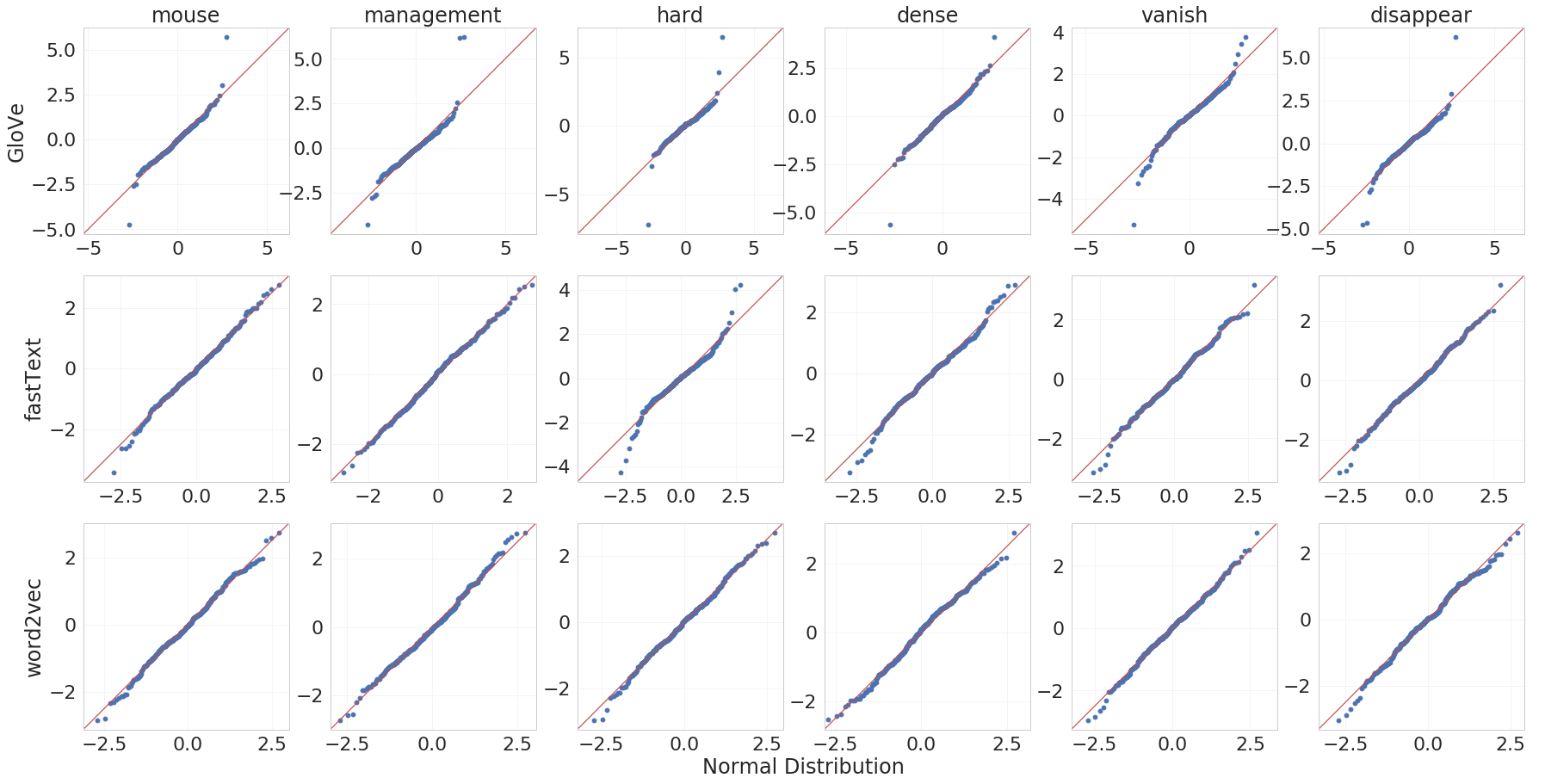}
	\caption{
	Q-Q plots comparing the theoretical quantiles of a standard normal distribution (horizontal axis) against the sample quantiles of standardised (Mean 0, SD 1) word vectors from three commonly used models:
	GloVe~\cite{Pennington2014}, fastText~\cite{Bojanowski2016}, and word2vec~\cite{Mikolov2013a,Mikolov2013b}.
	Perfect fit to the 45-degree reference line would indicate perfect normality.
	Note the pronounced discrepancy between the normal distribution and GloVe vectors due to the presence of heavy outliers.
	The discrepancy is relatively less pronounced for fastText vectors and much less pronounced for word2vec vectors.
	\Cref{fig:scatter} provides an alternative visualisation of the same phenomena.
	}
	\label{fig:qqplot}
\end{figure*}

The main point here is the following.
It is tempting to chose cosine similarity as the default and apply it everywhere regardless of the embedding type.
Sometimes, however, we should resist using what is convenient and instead use what is appropriate.
For example, if the samples corresponding to the marginals $P(X)$ and $P(Y)$ already look non-normal, then we conclude the joint $P(X, Y)$ cannot be a bivariate normal and the appropriateness of cosine similarity should be seriously questioned.
In some of these cases, using a rank-based coefficient as a similarity measure between word embeddings would be a much better alternative.
It will capture the association better, which could in turn lead to large improvements in performance on the downstream tasks.
In general, of course, even normal marginals do not imply a normal joint and care should be exercised either way; however we found the normality of marginals to be a good indication for cosine similarity within the scope of the present work.
In the next section we illustrate how the ideas discussed here can be applied in practice.

\section{Statistical Analysis of Word Embeddings: A Practical Example}
\label{sec:practice}

No matter how mysterious word vectors appear to be, just like any samples, they are subject to the full power of traditional statistical analysis.
As a concrete example, let's say we decided to use GloVe vectors \citep{Pennington2014}.
We treat each vector $\mbw_i$ as if it was a sample of 300 observations from some scalar random variable $W_i$.
We take a few hundred of these vectors, run a normality test such as Shapiro-Wilk \citep{ShapiroWilk1965} and find that the majority of them look non-normal ($p<0.05$).
As there is a considerable evidence against normality, we flag these vectors as `suspicious' and look at them closer.
We pick a few vectors and examine their histograms and Q-Q plots, seen in \Cref{fig:scatter} and \Cref{fig:qqplot} respectively;
the latter in particular is a statistical tool used to compare empirical and theoretical data distributions, and is explained further in the caption of \Cref{fig:qqplot}.
In both cases we observe that while the bulk of the distribution looks bell-shaped, we always get a couple of very prominent outliers.

Next, we can also visualise our word vectors in a way more directly relevant to the task at hand.
We take some pairs of words that are very similar (e.g. `vanish' and `disappear'), moderately similar (`hard' and `dense'),
and completely dissimilar (`mouse' and `management') and make the scatter plots for the corresponding pairs of word vectors.
These are also presented in \Cref{fig:scatter}.
We see that for similar pairs the relationship is almost linear; it becomes less linear as the similarity decreases, until we see a spherical blob (no relationship) for the most dissimilar pair.
However, we again face the presence of bivariate outliers that are too far away from the main bulk of points.

Given this evidence, which course of action shall we take?
Based on the presence of heavy outliers, we reject the normality of GloVe vectors and rule out the use of Pearson's $r$ and cosine similarity.
Instead we can use rank correlation coefficients, such as Spearman's $\rho$ or Kendall's $\tau$, as they offer more robustness to outliers.
Note that in this particular case, it may also be acceptable to winsorize (clip) the vectors and only then proceed with the standard Pearson's $r$.
We evaluate the proposed solution on word-level similarity tasks and observe good improvement in performance over cosine similarity, as seen in \Cref{tab:word-sim}.

Of course this exploration is in no way specific to GloVe vectors.
Note that from \Cref{fig:scatter} and \Cref{fig:qqplot}, we also see that word2vec vectors in particular tend to be much more normally distributed,
meaning that we don't find strong evidence against using Pearson correlation; this is again backed up by \Cref{tab:word-sim}.

This example helps illustrate that proper statistical analysis applied to existing textual embeddings is extremely powerful and comparatively less time-consuming than inventing new approaches.
Of course, this analysis can be made as fine-grained as desired.
Quite coarsely, we could have rejected the use of cosine similarity right after the Shapiro-Wilk test; on the other hand, we could have used even more different tests and visualisations.
The decision here rests with the practitioner and depends on the task and the domain.

\begin{table}[th!]
	\begin{tabular}{s@{\hskip 7pt}kj@{\hskip 8pt}s@{\hskip 8pt}ss@{\hskip 8pt}ss}
		\toprule
		\textbf{} & \textbf{task} & \textbf{N} & \textbf{V} & \textbf{COS} & \textbf{PRS} & \textbf{SPR} & \textbf{KEN} \\
		\midrule
		\toprule
		\textbf{\multirow{9}{*}{\rotatebox[origin=c]{90}{GloVe}}} & \textsc{yp-130} & .01 & = & 57.1 & 57.0 & 60.2 & 59.9 \\
		& \textsc{mturk-287} & .13 & = & 69.3 & 69.3 & 70.8 & 70.9 \\
		& \textsc{simlex-999} & .04 & R & 40.8 & 40.9 & 46.0 & 46.0 \\
		& \textsc{mc-30} & .10 & = & 78.6 & 79.2 & 77.0 & 77.4 \\
		& \textsc{simverb-3500} & .04 & R & 28.3 & 28.3 & 34.3 & 34.3 \\
		& \textsc{rg-65} & .14 & = & 76.2 & 75.9 & 71.0 & 71.1 \\
		& \textsc{ws-353-sim} & .06 & = & 80.3 & 80.2 & 80.1 & 80.1 \\
		& \textsc{verb-143} & .00 & = & 34.1 & 33.9 & 37.8 & 37.4 \\
		& \textsc{rw-stanford} & .16 & R & 46.2 & 46.2 & 52.8 & 52.9 \\
		\toprule
		\textbf{\multirow{9}{*}{\rotatebox[origin=c]{90}{fastText}}} & \textsc{yp-130} & .73 & = & 62.5 & 62.6 & 65.3 & 65.0 \\
		& \textsc{mturk-287} & .88 & = & 72.6 & 72.7 & 73.4 & 73.3 \\
		& \textsc{simlex-999} & .76 & = & 50.3 & 50.2 & 50.4 & 50.2 \\
		& \textsc{mc-30} & .90 & = & 85.2 & 85.2 & 84.6 & 84.5 \\
		& \textsc{simverb-3500} & .68 & = & 42.6 & 42.6 & 42.6 & 42.5 \\
		& \textsc{rg-65} & .90 & N & 85.9 & 85.8 & 83.9 & 84.1 \\
		& \textsc{ws-353-sim} & .84 & N & 84.0 & 83.8 & 82.4 & 82.2 \\
		& \textsc{verb-143} & .21 & = & 44.7 & 44.9 & 43.8 & 44.3 \\
		& \textsc{rw-stanford} & .80 & = & 59.5 & 59.4 & 59.0 & 58.9 \\
		\toprule
		\textbf{\multirow{9}{*}{\rotatebox[origin=c]{90}{word2vec}}} & \textsc{yp-130} & .95 & = & 55.9 & 56.1 & 55.0 & 54.7 \\
		& \textsc{mturk-287} & .94 & = & 68.4 & 68.3 & 67.1 & 67.2 \\
		& \textsc{simlex-999} & .94 & = & 44.2 & 44.2 & 43.9 & 44.0 \\
		& \textsc{mc-30} & .92 & = & 78.8 & 77.9 & 76.9 & 76.9 \\
		& \textsc{simverb-3500} & .96 & = & 36.4 & 36.4 & 36.0 & 36.0 \\
		& \textsc{rg-65} & .94 & = & 75.0 & 74.3 & 73.9 & 74.2 \\
		& \textsc{ws-353-sim} & .92 & N & 77.2 & 76.9 & 75.8 & 75.8 \\
		& \textsc{verb-143} & .98 & = & 49.7 & 50.1 & 48.9 & 49.0 \\
		& \textsc{rw-stanford} & .95 & N & 53.4 & 53.5 & 52.5 & 52.5 \\
		\toprule
	\end{tabular}
	\caption{Spearman's $\rho$ on word similarity tasks for combinations of word vectors and the following similarity metrics: cosine similarity (COS), Pearson's $r$ (PRS), Spearman's $\rho$ (SPR), and Kendall~$\tau$ (KEN). \textbf{N} indicates the proportion of sentence vectors in a task for which the null hypothesis of normality in a Shapiro-Wilk test was \emph{not} rejected at $\alpha=0.05$. The \textbf{V} column indicates the type of the best performing method: a rank-based correlation coefficient (R), a non-rank-based correlation or measure (N), or a tie (=). The winners in \textbf{V} were determined by comparing the top rank-based method for that vector/task combination with the top non-rank-based method. Winners were assigned only when the difference was statistically significant as determined by 95\% BCa confidence intervals.}
	\label{tab:word-sim}
\end{table}

\section{Experiments}
\label{sec:experiments}

To empirically validate the utility of the statistical framework presented in \Cref{sec:theory}, we run a set of evaluations on word- and sentence-level STS tasks.
In all experiments we rely on the following publicly available word embeddings: GloVe \citep{Pennington2014} trained on Common Crawl (840B tokens), fastText \citep{Bojanowski2016} trained on Common Crawl (600B tokens), and word2vec \citep{Mikolov2013a,Mikolov2013b} trained on Google News.
All the source code for our experiments is available on GitHub\footnote{\url{https://github.com/Babylonpartners/corrsim}}; in the case of the sentence-level tasks we rely also on the SentEval toolkit \citep{conneau2018senteval}.

First we consider a group of word-level similarity datasets that are commonly used as benchmarks in previous research: \textit{WS-353-SIM} \citep{Finkelstein2001}, \textit{YP-130} \citep{Yang2005}, \textit{SIMLEX-999} \citep{Hill2015}, \textit{SimVerb-3500} \citep{Gerz2016}, \textit{RW-STANFORD} \citep{Luong2013}, \textit{Verb-143} \citep{Baker2014}, \textit{MTurk-287} \citep{Radinsky2011}, \textit{MC-30} \citep{Miller1991}.
These datasets contain pairs of words and a human-annotated similarity score for each pair.
The success metric for the experiments is the Spearman correlation between the human-annotated similarity scores and the scores generated by the algorithm.
To avoid any confusion whatsoever, note that here Spearman correlation serves as an evaluation criterion; this is completely unrelated to using Spearman correlation as a similarity measure between word embeddings as proposed in \Cref{sec:theory}.
Bias-corrected and accelerated bootstrap \citep{Efron1987} 95\% confidence intervals were used to determine statistical significance.
We report the results for different combinations of word vectors and similarity measures in \Cref{tab:word-sim}.
The main takeaways from these experiments are the following:
\begin{itemize}
    \item There is no significant difference between the results obtained with cosine similarity and Pearson correlation.
    This is because empirically, the means across dimensions of these word vectors are approximately zero, in which case cosine similarity and Pearson correlation are approximately the same.
    \item Rank correlation coefficients tend to perform on par or better than cosine and Pearson on tasks and word vectors where there is a high proportion of non-normally distributed word vectors (over 90\%).
    This makes sense because it is precisely in the non-normal cases where Pearson correlation fails.
    \item When word vectors seem mostly normal, our analysis does not tell us definitively whether cosine similarity or rank correlation should perform better,
    and indeed we see that cosine and Pearson perform on par or better than Spearman and Kendall.
\end{itemize}

\begin{table}[th!]
	\centering
	\begin{tabular}{s@{\hskip 7pt}k@{\hskip 9pt}s@{\hskip 9pt}ss@{\hskip 9pt}ss@{\hskip 9pt}s}
		\toprule
		\textbf{} & \textbf{task} & \textbf{N} & \textbf{COS} & \textbf{PRS} & \textbf{SPR} & \textbf{KEN} & \textbf{APS}\\
		\midrule
		\midrule
		\multirow{5}{*}{\rotatebox[origin=c]{90}{\textbf{GloVe}}} & \textsc{STS12} & .01 & 52.1 & 52.0 & 53.4 & 52.6 & \textbf{53.8}\\
		& \textsc{STS13} & .00 & 49.6 & 49.6 & 56.2 & \textbf{56.7} & 55.9\\
		& \textsc{STS14} & .00 & 54.6 & 54.5 & \textbf{63.2} & 63.0 & 63.0\\
		& \textsc{STS15} & .00 & 56.1 & 56.0 & 64.5 & \textbf{65.3} & 64.2\\
		& \textsc{STS16} & .00 & 51.4 & 51.4 & 62.1 & \textbf{63.7} & 60.8\\
		\midrule
		\multirow{5}{*}{\rotatebox[origin=c]{90}{\textbf{fastText}}} & \textsc{STS12} & .01 & 58.3 & 58.3 & \textbf{60.2} & 59.0 & 58.4\\
		& \textsc{STS13} & .01 & 57.9 & 58.0 & 65.1 & \textbf{65.3} & 61.8\\
		& \textsc{STS14} & .00 & 64.9 & 65.0 & \textbf{70.1} & 69.6 & 68.5\\
		& \textsc{STS15} & .00 & 67.6 & 67.6 & 74.4 & \textbf{74.6} & 72.7\\
		& \textsc{STS16} & .00 & 64.3 & 64.3 & 73.0 & \textbf{73.5} & 70.7\\
		\midrule
		\multirow{5}{*}{\rotatebox[origin=c]{90}{\textbf{word2vec}}} & \textsc{STS12} & .95 & 51.6 & 51.6 & 51.7 & \textbf{53.1} & 45.3\\
		& \textsc{STS13} & .94 & 58.2 & \textbf{58.3} & 57.9 & 58.2 & 57.2\\
		& \textsc{STS14} & .96 & \textbf{65.6} & \textbf{65.6} & 65.5 & \textbf{65.6} & 64.1\\
		& \textsc{STS15} & .96 & 67.5 & 67.5 & 67.3 & \textbf{68.3} & 66.5\\
		& \textsc{STS16} & .96 & 64.7 & 64.7 & 64.6 & \textbf{65.6} & 63.9\\
		\toprule
	\end{tabular}
	\caption{Mean Pearson correlation on STS tasks for methods using combinations of word vectors and similarity metrics. All methods use averaged word vectors to represent sentences. The similarity measures are: cosine similarity (COS), Pearson's $r$ (PRS), Spearman's $\rho$ (SPR), Kendall~$\tau$ (KEN) and APSynP (APS). \textbf{N} indicates the proportion of sentence vectors in a task for which the null hypothesis of normality in a Shapiro-Wilk test was \emph{not} rejected at $\alpha=0.05$ }
	\label{tab:sts}
\end{table}

In the second set of experiments, we use the datasets from the sentence-level Semantic Textual Similarity shared task series 2012-2016~\citep{Agirre2012, Agirre2013a, Agirre2014, Agirre2015, Agirre2016, Cer2017}.
The success metric for these experiments is the Pearson correlation between the human-annotated sentence similarity scores and the scores generated by the algorithm.
Again, this use of Pearson correlation as an evaluation criterion is completely unrelated to its use as a similarity measure between sentence embeddings.
Note that the dataset for the STS13 SMT subtask is no longer publicly available, so the mean Pearson correlations reported in our experiments involving this task have been re-calculated accordingly.

For these experiments we use averaged word vectors as a sentence representation for various types of word vector, with similarity computed by the different correlation coefficients as well as cosine similarity and APSynP ~\cite{Santus2018}.
We report these results in \Cref{tab:sts}, and the full significance analysis for each subtask in \Cref{tab:cis}.
We also compare the top performing combination of averaged word vectors and correlation coefficient against several popular approaches from the literature that use cosine similarity:
BoW with ELMo embeddings \citep{Peters2018}, Skip-Thought \citep{Kiros2015}, InferSent \citep{Conneau2017}, Universal Sentence Encoder with DAN and Transformer \citep{USE2018}, and STN multitask embeddings \citep{Subramanian2018}.
These results are presented in \Cref{tab:sts-sota}.
Our observations for the sentence-level experiments are as follows:
\begin{itemize}
    \item The conclusions from the word-level tasks continue to hold and are even more pronounced: in particular, cosine and Pearson are essentially equivalent,
    and the increase in performance of rank-based correlation coefficients over cosine similarity on non-normal sentence vectors is quite dramatic.
    \item Averaged word vectors compared with rank correlation easily rival modern deep representations compared with cosine similarity.
\end{itemize}

\begin{table}[t]
	\centering
	\begin{tabular}{ksssss}
		\toprule
		\textbf{Approach~~~~~~~~STS} & \textbf{12~~} & \textbf{13~~} & \textbf{14~~} & \textbf{15~~} & \textbf{16~~} \\
		\midrule
		\midrule
		ELMo (BoW)        & 55~~          & 53~~          & 63~~          & 68~~          & 60~~          \\
		Skip-Thought      & 41~~          & 29~~          & 40~~          & 46~~          & 52~~          \\
		InferSent         & \textbf{61~~}          & 56~~          & 68~~          & 71~~          & 71~~          \\
		USE (DAN)         & 59~~          & 59~~          & 68~~          & 72~~          & 70~~          \\
		USE (Transformer) & \textbf{61~~}          & 64~~          & \textbf{71~~}          & 74~~          & \textbf{74~~}          \\
		STN (multitask)   & 60.6        & 54.7\textsuperscript{\textdagger}        & 65.8        & 74.2        & 66.4        \\
		\midrule
		fastText - COS    & 58.3 & 57.9 & 64.9 & 67.6 & 64.3 \\
		fastText - SPR     & 60.2        & 65.1        & 70.1        & 74.4        & 73.0          \\
		fastText - KEN    & 59.0          & \textbf{65.3}        & 69.6        & \textbf{74.6}        & 73.5       \\
		\toprule
	\end{tabular}
	\caption{
	Mean Pearson correlation on STS tasks for a variety of methods in the literature compared to averaged fastText
	vectors with different similarity metrics: cosine similarity (COS), Spearman's $\rho$ (SPR), and Kendall~$\tau$ (KEN).
	Values in bold indicate best results per task. Previous results are taken from \citet{perone2018evaluation}
	(only two significant figures provided) and \citet{Subramanian2018}.
	\textsuperscript{\textdagger}\ indicates the only STS13 result (to our knowledge) that includes the SMT subtask.
	}
	\label{tab:sts-sota}
\end{table}

Finally, the fraction of non-normal word vectors used in sentence-level tasks is consistent with the results reported for the word-level tasks in \Cref{tab:word-sim}.
However, we observe the following curious phenomenon for fastText.
While there is no evidence against normality for the majority of fastText vectors, perhaps surprisingly, when we average them to represent sentences, such sentence embeddings are almost entirely non-normal (\Cref{tab:sts}).
Empirically we observe that many high-frequency words or stopwords have prominently non-normal fastText vectors.
Although stopwords constitute only a small fraction of the entire vocabulary, they are very likely to occur in any given sentence, thus rendering most sentence embeddings non-normal as well.
While it's tempting to invoke the Central Limit Theorem (at least for longer sentences), under our formalism, averaging word vectors corresponds to averaging scalar random variables used to represent words, which are neither independent nor identically distributed.
In other words, there are no easy guarantees of normality for such sentence vectors.

\section{Discussion}
\label{sec:conclusion}

In this work, we investigate statistical correlation coefficients as measures for semantic textual similarity and make the following contributions:
\begin{itemize}
    \item We show that in practice, for commonly used word vectors, cosine similarity is equivalent to the Pearson
    correlation coefficient, motivating an alternative statistical view of word vectors as opposed to the geometric view, which is more prevalent in the literature.
    \item We illustrate via a concrete example the power and benefits of using elementary statistics to analyse word vectors.
    \item We characterise when Pearson correlation is applied inappropriately and show that these conditions hold for some
    word vectors but not others, providing a basis for deciding whether or not cosine similarity is a reasonable choice
    for measuring semantic similarity.
    \item We demonstrate that when Pearson correlation is not appropriate, non-parametric rank correlation
    coefficients, which are known to be more robust to various departures from normality, can be used as similarity measures to significantly
    improve performance on word- and sentence-level STS tasks.
    \item Finally, we show in particular that sentence representations consisting of averaged word vectors, when compared
    by rank correlation, can easily rival much more complicated representations compared by cosine similarity.
\end{itemize}

\begin{table*}[th!]
	\centering
	
	\begin{tabular}{s@{\hskip 7pt}k@{\hskip 9pt}ssk@{\hskip 9pt}ssk@{\hskip 9pt}ssk}
		\toprule
		&  & \multicolumn{3}{c}{\textbf{GloVe}} & \multicolumn{3}{c}{\textbf{fastText}} & \multicolumn{3}{c}{\textbf{word2vec}}\\
		\cline{3-11}
		&  & \textbf{SPR} & \textbf{COS} & \textbf{$\Delta$95\% BCa CI} & \textbf{SPR} & \textbf{COS} & \textbf{$\Delta$95\% BCa CI} & \textbf{SPR} & \textbf{COS} & \textbf{$\Delta$95\% BCa CI}\\
		\midrule
		\toprule
		\textbf{\multirow{5}{*}{\rotatebox[origin=c]{90}{STS12}}} & MSRpar & 35.90 & \textbf{42.55} & [-10.74, -2.52] & \textbf{39.66} & \textbf{40.39} & [-3.22, 1.80] & 38.79 & \textbf{39.72} & [-1.77, -0.16]\\
		& MSRvid & \textbf{68.80} & 66.21 & [1.31, 4.09] & \textbf{81.02} & 73.77 & [6.16, 8.53] & \textbf{77.88} & \textbf{78.11} & [-0.52, 0.06]\\
		& SMTeuroparl & \textbf{48.73} & \textbf{48.36} & [-5.26, 6.48] & 50.29 & \textbf{53.03} & [-5.41, -0.17] & \textbf{16.96} & 16.06 & [0.21, 1.34]\\
		& surprise.OnWN & \textbf{66.66} & 57.03 & [6.89, 12.76] & \textbf{73.15} & 68.92 & [2.19, 6.56] & \textbf{70.75} & \textbf{71.06} & [-0.73, 0.09]\\
		& surprise.SMTnews & \textbf{47.12} & \textbf{46.27} & [-4.27, 5.50] & \textbf{56.67} & \textbf{55.20} & [-2.50, 5.50] & \textbf{53.93} & \textbf{52.91} & [-0.13, 2.09]\\
		\midrule
		\textbf{\multirow{3}{*}{\rotatebox[origin=c]{90}{STS13}}} & FNWN & \textbf{43.21} & \textbf{38.21} & [-0.54, 10.24] & \textbf{49.40} & 39.83 & [2.74, 16.46] & \textbf{40.73} & \textbf{41.22} & [-2.07, 1.07]\\
		& headlines & \textbf{67.59} & 63.39 & [2.58, 5.89] & \textbf{71.53} & \textbf{70.83} & [-0.17, 1.58] & \textbf{65.48} & \textbf{65.22} & [-0.12, 0.66]\\
		& OnWN & \textbf{57.66} & 47.20 & [8.10, 13.02] & \textbf{74.33} & 63.03 & [9.27, 13.50] & 67.49 & \textbf{68.29} & [-1.29, -0.33]\\
		\midrule
		\textbf{\multirow{6}{*}{\rotatebox[origin=c]{90}{STS14}}} & deft-forum & \textbf{39.03} & 30.02 & [5.24, 13.52] & \textbf{46.20} & 40.19 & [2.88, 10.00] & \textbf{42.95} & \textbf{42.66} & [-0.43, 1.03]\\
		& deft-news & \textbf{68.99} & \textbf{64.95} & [-0.39, 8.72] & \textbf{73.08} & \textbf{71.15} & [-0.36, 4.39] & \textbf{67.33} & \textbf{67.28} & [-0.70, 0.91]\\
		& headlines & \textbf{61.87} & 58.67 & [1.15, 5.48] & \textbf{66.33} & \textbf{66.03} & [-0.68, 1.28] & \textbf{62.09} & \textbf{61.88} & [-0.22, 0.66]\\
		& images & \textbf{70.36} & 62.38 & [6.30, 10.00] & \textbf{80.51} & 71.45 & [7.44, 10.96] & 76.98 & \textbf{77.46} & [-0.89, -0.09]\\
		& OnWN & \textbf{67.45} & 57.71 & [7.89, 11.97] & \textbf{79.37} & 70.47 & [7.42, 10.50] & 74.69 & \textbf{75.12} & [-0.81, -0.08]\\
		& tweet-news & \textbf{71.23} & 53.87 & [13.98, 21.67] & \textbf{74.89} & 70.18 & [2.60, 7.21] & 68.78 & \textbf{69.26} & [-0.92, -0.01]\\
		\midrule
		\textbf{\multirow{5}{*}{\rotatebox[origin=c]{90}{STS15}}} & answers-forums & \textbf{50.25} & 36.66 & [10.18, 17.55] & \textbf{68.28} & 56.91 & [7.99, 15.23] & \textbf{53.74} & \textbf{53.95} & [-1.28, 0.86]\\
		& answers-students & \textbf{69.99} & 63.62 & [4.25, 9.59] & \textbf{73.95} & 71.81 & [0.69, 3.56] & \textbf{72.45} & \textbf{72.78} & [-0.70, 0.04]\\
		& belief & \textbf{58.77} & 44.78 & [10.11, 19.05] & \textbf{73.71} & 60.62 & [9.64, 19.50] & \textbf{61.73} & \textbf{61.89} & [-0.84, 0.46]\\
		& headlines & \textbf{69.61} & 66.21 & [1.65, 5.29] & \textbf{72.93} & \textbf{72.53} & [-0.40, 1.20] & \textbf{68.58} & \textbf{68.72} & [-0.48, 0.23]\\
		& images & \textbf{73.85} & 69.09 & [3.45, 6.29] & \textbf{83.18} & 76.12 & [5.76, 8.58] & \textbf{80.04} & \textbf{80.22} & [-0.55, 0.18]\\
		\midrule
		\textbf{\multirow{5}{*}{\rotatebox[origin=c]{90}{STS16}}} & answer-answer & \textbf{43.99} & 40.12 & [0.90, 7.36] & \textbf{54.51} & 45.13 & [5.14, 15.93] & \textbf{43.41} & \textbf{43.14} & [-1.03, 1.43]\\
		& headlines & \textbf{67.05} & 61.38 & [2.43, 9.44] & \textbf{71.00} & \textbf{70.37} & [-0.93, 2.13] & \textbf{66.55} & \textbf{66.64} & [-0.66, 0.51]\\
		& plagiarism & \textbf{72.25} & 54.61 & [12.69, 23.74] & \textbf{84.45} & 74.49 & [6.38, 14.81] & 75.21 & \textbf{76.46} & [-2.31, -0.37]\\
		& postediting & \textbf{69.03} & 53.88 & [12.01, 19.06] & \textbf{82.73} & 68.76 & [7.55, 22.96] & \textbf{73.87} & \textbf{73.35} & [-0.08, 1.21]\\
		& question-question & \textbf{58.32} & 47.21 & [7.02, 18.18] & \textbf{72.29} & 62.62 & [6.35, 13.64] & \textbf{63.94} & \textbf{63.74} & [-1.03, 1.38]\\
		\midrule
	\end{tabular}
	\caption{Pearson correlations between human sentence similarity score and a generated score. Generated scores were produced via measuring Spearman correlation (SPR), as explained in \Cref{sec:theory}, and cosine similarity (COS) between averaged word vectors. Values in bold represent the best result for a subtask given a set of word vectors, based on a 95\% BCa confidence interval~\cite{Efron1987} on the differences between the two correlations. In cases of no significant difference, both values are in bold.}
	\label{tab:cis}
\end{table*}

We hope that these contributions will inspire others to carefully investigate and understand alternative measures of similarity.
This is particularly important in the realm of sentence representations, where there are many more complex ways of
constructing sentence representations from word embeddings besides the simple averaging procedure tested here.
It is worth exploring whether a more subtle application of rank correlation
could help push these more complex sentence representations to even better performance on STS tasks.

A final and fascinating direction of future work is to explain the non-normality of certain types of word vectors (and in particular the presence of outliers) by analysing their training procedures.
Preliminary investigations suggest that unsupervised objectives based on the distributional hypothesis are probably not to blame, as word vectors trained without relying
on the distributional hypothesis, such as those of~\citet{Wieting2015}, still exhibit non-normality to some degree.
The actual causes remain to be determined.
We believe that understanding the reasons for these empirically-observed characteristics of textual embeddings would be
a significant step forwards in our overall understanding of these crucial building blocks for data-driven natural language processing.

\section*{Acknowledgements}
\label{sec:acknowledgements}

We would like to thank Dan Busbridge and the three anonymous reviewers for their useful feedback and suggestions.

\pagebreak
\bibliography{naaclhlt2019}
\bibliographystyle{acl_natbib}

\end{document}